\documentclass[runningheads]{llncs}
\usepackage[pdftex]{graphicx}
\usepackage[latin1]{inputenc}
\usepackage[T1]{fontenc}
\usepackage{amssymb,amsmath,array}
\usepackage{cleveref} 
\usepackage{caption}
\usepackage{enumitem}
\usepackage[autostyle=true]{csquotes}
\usepackage{subfig}
\usepackage[export]{adjustbox} 
\usepackage{siunitx}
\usepackage{float}
\usepackage{url}

\crefname{section}{Sec.}{Secs.}
\crefname{subsection}{Sec.}{Secs.}
\crefname{subsubsection}{Sec.}{Secs.}
\crefname{table}{Tab.}{Tabs.}
\crefname{figure}{Fig.}{Figs.}
\crefname{equation}{Eq.}{Eq.}
\Crefname{equation}{Eq.}{Eq.}
\crefname{algorithm}{Alg.}{Algs.}
\crefname{appendix}{App.}{Apps.}
\newcommand{\vmu}{\vec{\mu}}
\newcommand{\vx}{\vec{x}}

\newcommand{\vd}{\vec{d}}

\newcommand{\wrt}{w.r.t.\ }
\newcommand{\mSigma}{\Sigma}
\newcommand{\mP}{\textbf{P}}
\newcommand{\mD}{\textbf{D}}
\newcommand{\mX}{\textbf{X}}
\newcommand{\Cov}{\text{Cov}}
\newcommand{\E}{\mathbb{E}}

\let\oldsqrt\sqrt
\def\sqrt{\mathpalette\DHLhksqrt}
\def\DHLhksqrt#1#2{%
  \setbox0=\hbox{$#1\oldsqrt{#2\,}$}\dimen0=\ht0
  \advance\dimen0-0.2\ht0
  \setbox2=\hbox{\vrule height\ht0 depth -\dimen0}%
  {\box0\lower0.4pt\box2}}

\begin{document}
\title{A Rigorous Link Between Self-Organizing Maps and Gaussian Mixture Models}
\titlerunning{A Rigorous Link Between SOMs and GMMs} 
\author{Alexander Gepperth \and Benedikt Pfülb}
\authorrunning{A.\ Gepperth and B.\ Pfülb} 
\institute{University of Applied Sciences Fulda, Leipzigerstr.123, 36037 Fulda, Germany
\email{\{alexander.gepperth,benedikt.pfuelb\}@cs.hs-fulda.de}\\
\url{www.hs-fulda.de}
}
\maketitle              
%
\begin{abstract}
This work presents a mathematical treatment of the relation between Self-Organizing Maps (SOMs) and Gaussian Mixture Models (GMMs).
We show that energy-based SOM models can be interpreted as performing gradient descent, minimizing an approximation to the GMM log-likelihood that is particularly valid for high data dimensionalities.
The SOM-like decrease of the neighborhood radius can be understood as an annealing procedure ensuring that gradient descent does not get stuck in undesirable local minima.
This link allows to treat SOMs as generative probabilistic models, giving a formal justification for using SOMs, e.g., to detect outliers, or for sampling.
\end{abstract}

\section{Introduction}
This theoretical work is set in the context of unsupervised clustering and density estimation methods and establishes a mathematical link between two important representatives: Self-Organizing Maps (SOMs, \cite{Cottrell1994,Kohonen1990}) and Gaussian Mixture Models (GMMs, \cite{Dempster1977}), both of which have a long history in machine learning.
There are significant overlaps between SOMs and GMMs, and both models have been used for data visualization and outlier detection.
They are both based on Euclidean distances and model data distributions by prototypes or centroids.
At the same time, there are some differences:
GMMs, as fully generative models with a clear probabilistic interpretation, can additionally be used for sampling purposes.
Typically, GMMs are trained batch-wise, repeatedly processing all available data in successive iterations of the Expectation-Maximization (EM) algorithm.
In contrast, SOMs are trained online, processing one sample at a time.
The training of GMMs is based on a loss function, usually referred to as \textit{incomplete-data log-likelihood} or just log-likelihood.
Training by Stochastic Gradient Descent (SGD) is possible, as well, although few authors have explored this~\cite{Hosseini2015}.
SOMs are not based on a loss function, but there are model extensions~\cite{Gepperth2019,Heskes1999} that propose a simple loss function at the expense of very slight differences in model equations.
Lastly, GMMs have a simple probabilistic interpretation as they attempt to model the density of observed data points.
For this reason, GMMs may be used for outlier detection, clustering and, most importantly, sampling.
In contrast to that, SOMs are typically restricted to clustering and visualization due to the topological organization of prototypes, which does not apply to GMMs.
\subsection{Problem Statement}
SOMs are simple to use, implement and visualize, and, despite the absence of theoretical guarantees, have a very robust training convergence.
However, their interpretation remains unclear.
This particularly concerns the probabilistic meaning of input-prototype distances.
Different authors propose using the Best-Matching Unit (BMU) position only, while others make use of the associated input-prototype distance, or even the combination of all distances~\cite{Hecht2015}.
Having a clear interpretation of these quantities would help researchers tremendously when interpreting trained SOMs.
The question whether SOMs actually perform density estimation is important for justifying outlier detection or clustering applications.
Last but not least, a probabilistic interpretation of SOMs, preferably a simple one in terms of the well-known GMMs, would help researchers to understand how sampling from SOMs can be performed.
\subsection{Results and Contribution}
This article aims at explaining SOM training as Stochastic Gradient Descent (SGD) using an energy function that is a particular approximation to the GMM log-likelihood.
SOM training is shown to be an approximation to training GMMs with tied, spherical covariance matrices where constant factors have been discarded from the probability computations.
This identification allows to interpret SOMs in a probabilistic manner, particularly for:
\begin{itemize}[leftmargin=*]
  \setlength\itemsep{0em}
  \item outlier detection (not only the position of the BMU can be taken into account, but also the associated input-prototype distance since it has a probabilistic interpretation) and
  \item sampling (understanding what SOM prototypes actually represent, it is possible to generate new samples from SOMs with the knowledge that this is actually sanctioned by theory)
\end{itemize}
\subsection{Related Work}
Several authors have attempted to establish a link between SOMs and GMMs.
In \cite{Heskes2001}, an EM algorithm for SOMs is given, emphasizing the close links between both models.
Verbeek et al.~\cite{Verbeek2005} emphasizes that GMMs are regularized to show a SOM-like behavior of self-organization.
A similar idea of component averaging to obtain SOM-like normal ordering and thus improved convergence was previously demonstrated in \cite{Ormoneit1998}.
An energy-based extension of SOMs suggesting a close relationship to GMMs is given in \cite{Heskes1999}, with an improved version described in \cite{Gepperth2019}.
So far, no scientific work has tried to explain SOMs as an approximation to GMMs in a way that is comparable to this work.
\section{Main Proof}
\begin{figure}[ht]
  \centering
  \vspace{-1em}
  \includegraphics[width=\textwidth]{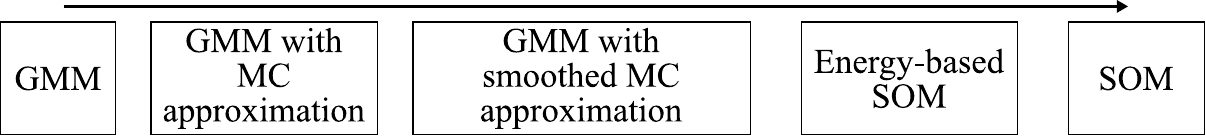}
  \caption{Outline of main proof.\label{fig:1}}
\end{figure}

The general outline of proof is depicted in \cref{fig:1}.
We start with a description of GMMs in \cref{gmm:full}.
Subsequently, a transition from exact GMMs to the popular Max-Component (MC) approximation of its loss is described in \cref{sec:max-comp}.
Then, we propose a SOM-inspired annealing procedure for performing optimization of approximate GMMs in \cref{gmm:ann} and explain its function in the context of SGD.
Finally, we show that this annealed training procedure is equivalent to training energy-based SOMs in \cref{som}, which are a faithful approximation of the original SOM model, outlined in \cref{sec:som}.
\subsection{Default GMM Model, Notation and GMM Training}\label{gmm:full}
Gaussian Mixture Models (GMMs) are probabilistic latent-variable models that aim to explain the distribution of observed data samples $X$=$\{\vec{x}_n\}$.
It is assumed that samples generated from a known parametric distribution, which depends on parameters $\theta$ and unobservable latent variables $Z$=$\{\vec{z}_n$\,$\in$\,$\mathbb{R}^K\}$, $z_{nk}$\,$\in$\,$\{0,1\}$ and $\sum_k z_{nk}$\,$=$\,$1$.
The \textit{complete-data probability} reads
\begin{align}
p(X,Z) = \prod_n^N\prod_k^K \big[p_k(\vec{x}_n)\big]^{\pi_k z_{nk}}
\end{align}
where mixture components $p_k(\vec{x}_n)$ are modeled as multi-variate Gaussians, and whose parameters are the centroids $\vmu_k$ and the positive-definite covariance matrices $\mSigma_k$ (both omitted from the notation for conciseness).
$p_k(\vec{x}_n)$ represents the probability of observing the data vector $\vec{x}_n$, if sampled from mixture component $k$.
The number of Gaussian mixture components $K$ is a free parameter of the model, and the component weights $\pi_k$ must have a sum of $1$.
Since the latent variables are not observable, it makes sense to marginalize them out by summing over the discrete set of all possible values $\mathcal{Z}$\,$=$\,$\{\vec{e}_j| j$\,$=$\,$1\dots K\}$, giving
\begin{align}
p(X) = \prod_n \sum_{\vec{z}_n\in\mathcal{Z}}\prod_k p_k(\vec{x}_n)^{\pi_k z_{nk}}\text{.}
\end{align}
Taking the logarithm and normalizing by the number of samples ($N$) provides the \textit{incomplete-data log-likelihood}, which contains only observable quantities and is, therefore, a suitable starting point for optimization:
\begin{equation}
\mathcal{L} = \frac{1}{N}\sum_{n=1}^N \log \sum_k \pi_k p_k(\vec{x}_n)\text{.}
\label{eqn:loglik}
\end{equation}
\subsection*{Concise Problem Statement}
When training GMMs, one aims at finding parameters $\vec{\mu}_k$, $\Sigma_k$ that (locally) maximize \cref{eqn:loglik}.
This is usually performed by using a procedure called Expectation-Maximization (EM, \cite{Dempster1977,Hartley1958}) which is applicable to many latent-data (mixture) models.
Of course, a principled alternative to EM is an approach purely based on batches or SGD, the latter being an approximation justified by the Robbins-Monro procedure \cite{Robbins1951}.
In this article, we will investigate how SGD optimization of \cref{eqn:loglik} can be related to the training of SOMs.
\subsection*{Respecting GMM Constraints in SGD}\label{sec:constr}
GMMs impose the following constraints on the parameters $\pi_k$, $\vmu_k$ and $\mSigma_k$: 
\begin{itemize}[leftmargin=*]
  \setlength\itemsep{0em}
  \item weights must be normalized: $\sum_k \pi_k$\,$=$\,$1$
  \item covariance matrices must be positive-definite: $\vx^T\mSigma_k\vx \ge 0\,\forall\,k,\vx$
\end{itemize}
The first constraint can be enforced after each gradient decent step by setting $\pi_k$\,$\rightarrow$\,$\frac{\pi_k}{\sum_j \pi_j}$.
For the second constraint, we consider diagonal covariance matrices only, which is sufficient for establishing a link to SOMs.
A simple strategy in this setting is to re-parameterize covariance matrices $\mSigma_k$ by their inverse (denoted as precision matrices) $\mP_k$\,$=$\,$\mSigma^{-1}$.
We then re-write this as $\mP_k$\,$=$\,$\mD_k\mD_k$, which ensures positive-definiteness of $\mP$, $\mSigma$. 
The diagonal entries $\vec{s}_k$ of $\mSigma_k$ can thus be re-written as $s_{ki}$\,$=$\,$d_{ki}^{-2}$, whereas $\vd_k$ are the diagonal entries of $\mD_k$. 
\subsection{Max-Component Approximation}\label{sec:max-comp}
In \cref{eqn:loglik}, we observe that the component weights $\pi_k$ and the conditional probabilities $p(\vec{x})$ are positive by definition.
It is, therefore, evident that any single component of the inner sum over the components $k$ is a lower bound of the entire inner sum.
The largest of these $K$ lower bounds is given by the maximum of the components, so that it results in
\begin{equation}
\begin{split}
  \mathcal{L} & = \frac{1}{N}\sum_{n=1}^N \log \sum_k \pi_k p(\vec{x}_n ) \le \hat{\mathcal{L}} = \frac{1}{N}\sum_n \log \text{max}_k \big(\pi_k p(\vec{x}_n)  \big) \\
              & = \frac{1}{N}\sum_n  \text{max}_k \log \big(\pi_k p(\vec{x}_n) \big).
\end{split}
\label{eqn:max}
\end{equation}
\Cref{eqn:max} displays what we refer to as \textit{Max-Component approximation} to the log-likelihood.
Since $\hat{\mathcal{L}}$\,$\le$\,$\mathcal{L}$, we can increase $\mathcal{L}$ by maximizing $\hat{\mathcal{L}}$.
The advantage of $\hat{\mathcal{L}}$ is that it is not affected by numerical instabilities the way $\mathcal{L}$ is.
Moreover, it breaks the symmetry between mixture components, thus, avoiding degenerate local optima during early training.
Apart from facilitating the relation to SOMs, this is an interesting idea in its own right, which was first proposed in \cite{Dognin2009}.
\subsection*{Undesirable Local Optima}
GMMs are usually trained using EM after a k-means initialization of the centroids.
Since this work explores the relation to SOMs, which are mainly trained from scratch, we investigate SGD-based training of GMMs without k-means initialization.
A major problem in this setting are undesirable local optima, both for the full log-likelihood $\mathcal{L}$ and its approximation $\hat{\mathcal{L}}$.
To show this, we parameterize the component probabilities by the precision matrices $\mP_k$\,$=$\,$\mSigma_k^{-1}$ and compute
\begin{gather}
\begin{aligned}
  \frac{\partial \mathcal{L}}{\partial \vmu_{k}} & = \E_n\left[  \mP_k\left(\vx_{n}-\vmu_{k}\right)\gamma_{nk}\right]                             \\[.5em]
  \frac{\partial \mathcal{L}}{\partial \mP_{k}}  & = \E_n\left[ \left( (\mP_k)^{-1} - (\vx_n - \vmu_k)(\vx_n - \vmu_k)^T\right)\gamma_{nk}\right] \\[.5em]
  \frac{\partial \mathcal{L}}{\partial \pi_{k}}  & = \pi_k^{-1}\E_n\left[\gamma_{nk}\right]
\end{aligned}
\label{eqn:mcgrads}
\raisetag{20pt}
\end{gather}
whereas $\gamma_{nk}\in[0,1]$ denote standard GMM responsibilities given by
\begin{align}
\gamma_{nk} = \frac{\pi_k p_k(\vx_n)}{\sum_k \pi_kp_k(\vx_n)}.
\end{align}
\par\noindent\textbf{Degenerate Solution}
This solution universally occurs when optimizing $\mathcal{L}$ by SGD, and represents an obstacle for naive SGD.
All components have the same weight, centroid and covariance matrix: $\pi_{k}$\,$\approx$\,$\frac{1}{K}$, $\vmu_{k}$\,$=$\,$\E[\mX]$, \mbox{$\mSigma_{k}$\,$=$\,$\Cov(\mX)$ $\forall$\,$k$}.
Since the responsibilities are now uniformly $1/K$, it results from \cref{eqn:mcgrads} that all gradients vanish.
This effect is avoided by $\hat{\mathcal{L}}$ as only a subset of components is updated by SGD, which breaks the symmetry of the degenerate solution.
\par\noindent\textbf{Single/Sparse-Component Solution}
Optimizing $\hat{\mathcal{L}}$ by SGD, however, leads to another class of unwanted local optima:
A single component $k^*$ has a weight close to $1$, with its centroid and covariance matrix being given by the mean and covariance of the data: $\pi_{k^*}$\,$\approx$\,$1$, $\vmu_{k^*}$\,$=$\,$\E[\mX]$, $\mSigma_{k^*}$\,$=$\,$\Cov(\mX)$.
For $\hat{\mathcal{L}}$, the gradients in \cref{eqn:mcgrads} stay the same except for $\gamma_{nk}$\,$=$\,$\delta_{nk^*}$ from which we conclude that the gradient \wrt $\mP_{k}$ and $\vmu_{k}$ vanishes $\forall k$.
The gradient \wrt $\pi_k$ does not vanish, but is $\delta_{kk^*}$, which disappears after enforcing the normalization constraint (see \cref{sec:constr}).
A variant is the sparse-component solution where only a few components have non-zero weights, so that the gradients vanish for the same reasons.
\subsection{Annealing Procedure}\label{sec:sgd-reg}\label{gmm:ann}
A simple SOM-inspired approach to avoid these undesirable solutions is to punish their characteristic response patterns by an appropriate modification of the (approximate) loss function that is maximized, i.e., $\hat{\mathcal{L}}$.
We introduce what we call \textit{smoothed Max-Component log-likelihood} $\hat{\mathcal{L}}^\sigma$, inspired by SOM training:
\begin{equation}
\begin{split}
  \hat{\mathcal{L}}^\sigma & = \frac{1}{N}\sum_n \text{max}_k \left( \sum_j  \vec{g}_{kj} \log \Big(\pi_j {p}(\vec{x}_n)\Big)\right).
\end{split}
\label{eqn:conv}
\end{equation}
Here, we assign a normalized coefficient vector $\vec{g}_k$ to each Gaussian mixture component $k$.
The entries of $\vec{g}_k$ are computed in the following way:
\begin{itemize}[leftmargin=*]
  \setlength\itemsep{0em}
  \item Assume that the $K$ Gaussian components are arranged on a 1D grid of dimensions $(1,K)$ or on a 2D grid of dimensions $(\sqrt{K},\sqrt{K})$.
        As a result, each linear component index $k$ has a unique associated 1D or 2D coordinate $\vec{c}(k)$.
  \item Assume that the vector $\vec{g}_k$ of length $K$ is actually representing a 1D structure of dimension $(1,K)$ or a 2D structure of dimension $(\sqrt{K},\sqrt{K})$.
        Each linear vector index $j$ in $\vec{g}_k$ has a unique associated 1D or 2D coordinate $\vec{c}(j)$.
  \item The entries of the vector $\vec{g}_k$ are computed as
        \begin{equation}
         g_{kj} = \exp\left(-\frac{\big(\vec{c}(j)-\vec{c}(k)\big)^2} {2\sigma^2}\right)
        \label{eqn:gauss}
        \end{equation}
\end{itemize}
and subsequently normalized to have a unit sum.
Essentially, \cref{eqn:conv} represents a convolution of $\log \pi_k p_k(\vec{x})$, arranged on a periodic 2D grid with a Gaussian convolution filter, resulting in a smoothing operation.
The 2D variance $\sigma$ in \cref{eqn:gauss} is a parameter that must be set as a function of the grid size so that Gaussians are neither homogeneous, nor delta peaks.
Hence, the loss function in \cref{eqn:conv} is maximized if the log probabilities follow an uni-modal Gaussian profile of variance $\sigma$, whereas single-component solutions are punished.
\par
It is trivial to see that the annealed loss function in \cref{eqn:conv} reverts to the non-annealed form \cref{eqn:max} in the limit where $\sigma$\,$\rightarrow$\,$0$.
This is due to the fact that vectors $\vec{g}_k$ approach Kronecker deltas in this case with only a single entry of value $1$.
Thereby, the inner sum in \cref{eqn:conv} is removed.
By making $\sigma$ time-dependent in a SOM-like manner, starting at a value of $\sigma(t_0)$\,$\equiv$\,$\sigma_0$ and then reducing it to a small final value $\sigma(t_{\infty})$\,$\equiv$\,$\sigma_{\infty}$.
The result is a smooth transition from the annealed loss function \cref{eqn:conv} to the original max-component log-likelihood \cref{eqn:max}.
Time dependency of $\sigma(t)$ can be chosen to be:
\begin{equation}\label{eqn:ann}
\sigma(t)=\left\{
  \begin{array}{cc}
          \sigma_0        &       t < t_0        \\
       \sigma_{\infty}    &    t > t_{\infty}    \\
    \sigma_0\exp(-\tau t) & t_0 < t < t_{\infty}
  \end{array}
\right.
\end{equation}
where the time constant in the exponential is chosen as $\tau$\,$=$\,$\log \frac{\sigma_0-\sigma_{\infty}}{t_{\infty}-t_0} $ to ensure a smooth transition.
This is quite common while training SOMs where the neighborhood radius is similarly decreased.
\subsection{Link to Energy-based SOM Models}\label{som}
The standard Self-Organizing Map (SOM) has no energy function that is minimized.
However, some modifications (see \cite{Gepperth2019,Heskes1999}) have been proposed to ensure the existence of a $C^\infty$ energy function.
These energy-based SOM models reproduce all features of the original model and use a learning rule that the original SOM algorithm is actually approximating very closely.
In the notation of this article, SOMs model the data through $K$ prototypes $\vec{\mu}_k$ and $K$ neighborhood functions $\vec{g}_k$ defined on a periodic 2D grid.
Their energy function is written as
\begin{equation}
  \mathcal{L}_{\textit{SOM}} = \frac{1}{N}\sum_n \text{min}_k \sum_j g_{kj}\|\vec{x}_n -\vec{\mu}_j\|^2\text{,}
  \label{eqn:som}
\end{equation}
whose optimization by SGD initiates the learning rule for energy-based SOMs:
\begin{equation}
  \vec{\mu}_k(t+1) = \vec{\mu}_k + \epsilon g_{ki^*} (\vec{x}-\vec{\mu}_k)
\end{equation}
with the Best-Matching Unit (BMU) having index $i^*$.
In contrast to the standard SOM model the BMU is determined as $i^*$\,$=$\,$\text{argmax}_i \sum_j g_{ik}||\vec{x}-\vec{\mu}_i||$.
The link to the standard SOM model is the observation that for small values of the neighborhood radius $\sigma(t)$ the convolution vanishes and the original SOM learning rule is recovered.
This is typically the case after the model has initially converged (sometimes referred to as \enquote{normal ordering}).
\subsection{Equivalence to SOMs}\label{sec:som}
When writing out $\log \pi_k p_k(\vx)$\,$=$\,$-\sum_j \frac{d_kj^2}{2} \left( \vx_j-\vmu_{kj}\right)$, tying the variances so that $\vd_{kj}$\,$=$\,$d\,\forall\,j$ and fixing the weights
to $\pi_k$\,$=$\,$\frac{1}{K}$ in \cref{eqn:conv} we find that the energy function \cref{eqn:som} becomes
\begin{align}
  \hat{\mathcal{L}}^\sigma & = \frac{1}{N}\sum_n \text{max}_k \left( \sum_j  {g}_{kj} \Big(-\log K - \frac{d^2}{2}||\vx_n-\vmu_j||\Big)\right) \\
                           & = \frac{d^2}{2N}\sum_n \text{min}_k \sum_j  {g}_{kj}  ||\vx_n-\vmu_j||+\text{const}\text{.}
\end{align}
In fact \cref{eqn:conv} is identical to \cref{eqn:som}, except for a constant factor that can be discarded and a scaling factor defined by the common tied precision $d^2$.
The minus sign just converts the max into a min operation, as distances and precisions are positive.
The annealing procedure of \cref {eqn:ann} is identical to the method for reducing the neighborhood radius during SOM training as well.
\par
Energy-based SOMs are a particular formulation (tied weights, constant spherical variances) of GMMs which are approximated by a commonly accepted method.
Training energy-based SOMs in the traditional way results in the optimization of GMMs by SGD, where training procedures are, again, identical.
\section{Experiments}
\begin{figure}[htb!]
\vspace{-1.em}
\centering
  \subfloat[Ensemble of \enquote{neighborhood functions} $g_{kj}$]{\includegraphics[width=\linewidth]{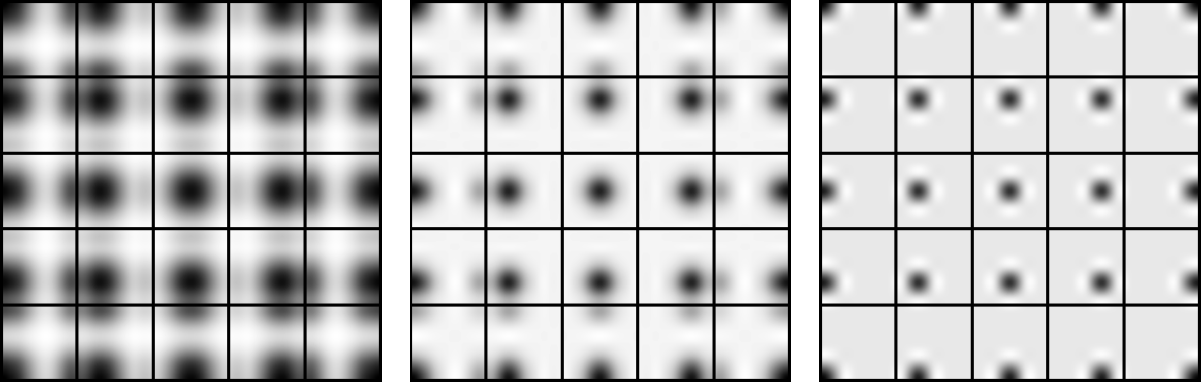}\label{fig:iter2}}
  \\
  \subfloat[Centroids $\mu_k$]{\includegraphics[width=\linewidth]{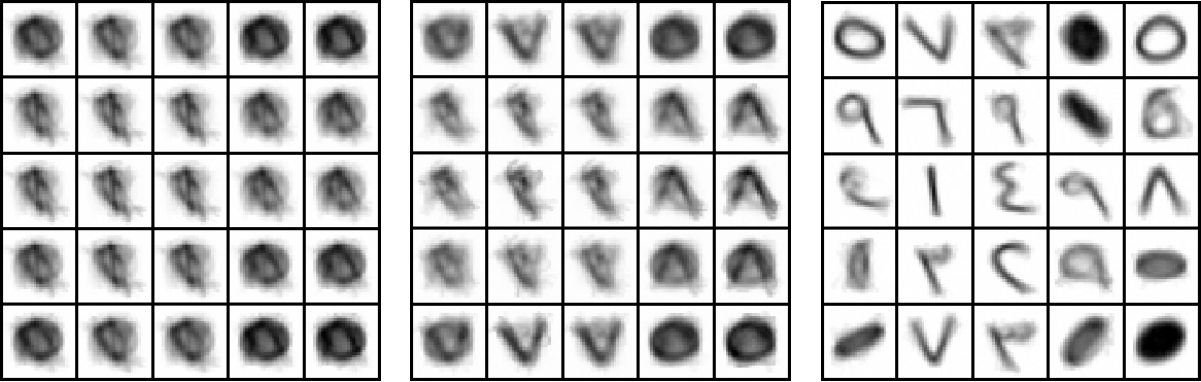}\label{fig:iter22}}
  \\
  \subfloat[Trend of $\sigma(t)$ and $\epsilon(t)$]{\includegraphics[width=.96\textwidth]{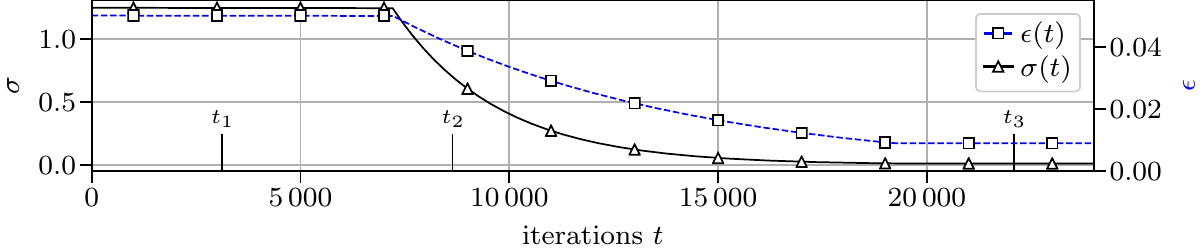}\label{fig:all}
  }
  \caption{\label{fig:conv_mask}
    Visualization of GMM evolution at three points in time during training (from left to right: iterations $t_1$\,$=$\,$3\,120$, $t_2$\,$=$\,$8\,640$ and $t_3$\,$=$\,\num{22080}).
  }
\end{figure}

This section presents a simple proof-of-concept that the described SGD-based training scheme is indeed practical.
It is not meant to be an exhaustive empirical proof and is, therefore, just conducted on a common image dataset.
For this experiment, we use the MADBase dataset~\cite{MADBase} containing \num{60000} grayscale images of handwritten arabic digits in a resolution of $28$\,$\times$\,$28$ pixels.
The number of training iterations $T$\,=\,$24\,000$ with a batch size of $1$, as it is common for SOMs.
We use a GMM with $K$\,$=$\,$25$ components, whose centroids are initialized to small random values, whereas weights are initially equiprobable and precisions are uniformly set to $d^2$\,=\,$5$ (we found that precisions should initially be as large as possible).
We set $t_0$\,$=$\,$0.3\,T$ and stop at $t_\infty$\,$=$\,$0.8\,T$ (proportional to the maximum number of iterations).
$\sigma_0$ starts out at $1.2$ (proportional to the map size) and is reduced to a value of $0.01$.
The learning rate is similarly decayed to speed up convergence, although this is not a requirement, with $\epsilon_0$\,=\,$0.05$ and $\epsilon_\infty$\,=\,$0.009$.
In \cref{fig:conv_mask}, three states of the trained GMM are depicted (at iteration $t_1$\,$=$\,$3120$, $t_2$\,$=$\,$8\,640$ and $t_3$\,$=$\,\num{22080}).
\captionsetup[subfigure]{labelformat=empty}
\par
We observe from the development of the prototypes that GMM training converges, and that the learned centroids are representing the dataset well.
It can also be seen that prototypes are initially blurred and get refined over time, which resembles SOMs.
No undesired local optima were encountered when this experiment was repeated $100$ times.
\section{Discussion}
\par\noindent\textbf{Approximations}
The approximations on the way from GMMs to SOMs are \cref{eqn:max} and the approximations made by energy-based SOM models (see \cite{Heskes2001} for a discussion). 
It is shown in \cite{Dognin2009} that the quality of the first approximation is excellent, since inter-cluster distances tend to be large, and it is more probable that a single component can explain the data.
\par\noindent\textbf{Consequences}
The identification of SOMs as special approximation of GMMs allows for the performance of typical GMM functions (sampling, outlier detection) with trained SOMs.
The basic quantity to be considered here is the input-prototype distance of the Best-Matching Unit (BMU) since it corresponds to a log probability.
In particular, the following consequences were to be expected:
\begin{itemize}[leftmargin=*]
  \setlength\itemsep{0em}
  \item \textbf{Outlier Detection}
        The value of the smallest input-prototype distance is the relevant one, as it represents $\hat{\mathcal{L}}$ for a single sample, which in turn approximates the incomplete-data log-likelihood $\mathcal {L}$.
        In practice, it can be advantageous to average $\hat{\mathcal{L}}$ over several samples to be robust against noise.
  \item \textbf{Clustering}
        SOM prototypes should be viewed as cluster centers, and inputs can be assigned to the prototype with the smallest input-prototype distance.
  \item \textbf{Sampling}
        Sampling from SOMs should be performed in the same way as from GMMs.
        Thus, in order to create a sample, a random prototype has to be selected first (since weights are tied no multinomials are needed here).
        Second, a sample from a Gaussian distribution with precision 2 and centered on the selected prototype along all axes needs to be drawn.
\end{itemize}

\subsection{Summary and Conclusion}
To our knowledge, this is the first time that a rigorous link between SOMs and GMMs has been established, based on a comparison of loss/energy functions.
It is thus shown that SOMs actually implement an annealing-type approximation to the full GMM model with fixed component weights and tied diagonal variances.
To give more weight to the mathematical proof, we validate the SGD-based approach to optimize GMMs in practice.
\clearpage
{\bibliographystyle{splncs04}
\bibliography{icann2020}}
Preprint Version. Accepted at International Conference on Artificial Neural Networks (ICANN) 2020.
\end{document}